\documentclass{article}

\usepackage{arxiv}

\usepackage[utf8]{inputenc} 
\usepackage[T1]{fontenc}    
\usepackage{hyperref}       
\usepackage{url}            
\usepackage{booktabs}       
\usepackage{amsfonts}       
\usepackage{nicefrac}       
\usepackage{microtype}      
\usepackage{lipsum}		
\usepackage{graphicx}
\usepackage{natbib}
\usepackage{doi}
\usepackage{multirow}
\usepackage{amsmath}
\usepackage{amssymb}
\usepackage{subcaption}
\usepackage{pgfplots}
\pgfplotsset{compat=1.18}
\usepackage{tikz}

\title{Combining Language and Topic Models for Hierarchical Text Classification}

\author{ \href{https://orcid.org/0000-0001-6782-2381}{\includegraphics[scale=0.06]{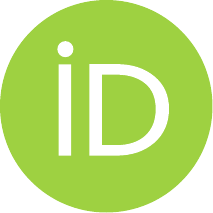}\hspace{1mm}Jaco du Toit} \\
	Computer Science Division, Mathematical Sciences Department\\
	Stellenbosch University\\
	Stellenbosch, South Africa \\
	\texttt{jacowdutoit11@gmail.com} \\
	\And
	\href{https://orcid.org/0000-0003-1957-3979}{\includegraphics[scale=0.06]{orcid.pdf}\hspace{1mm}Marcel Dunaiski} \\
	Computer Science Division, Mathematical Sciences Department\\
	Stellenbosch University\\
	Stellenbosch, South Africa \\
	\texttt{marceldunaiski@sun.ac.za} \\
}

\renewcommand{\shorttitle}{Combining Language and Topic Models for Hierarchical Text Classification}

\hypersetup{
pdftitle={Combining Language and Topic Models for Hierarchical Text Classification},
pdfauthor={Marcel Dunaiski},
pdfkeywords={Hierarchical Text Classification, Large Language Models, Topic Models},
}

\begin{document}
\maketitle

\begin{abstract}
Hierarchical text classification (HTC) is a natural language processing task which has the objective of categorising text documents into a set of classes from a predefined structured class hierarchy. Recent HTC approaches use various techniques to incorporate the hierarchical class structure information with the natural language understanding capabilities of pre-trained language models (PLMs) to improve classification performance. Furthermore, using topic models along with PLMs to extract features from text documents has been shown to be an effective approach for multi-label text classification tasks. The rationale behind the combination of these feature extractor models is that the PLM captures the finer-grained contextual and semantic information while the topic model obtains high-level representations which consider the corpus of documents as a whole. In this paper, we use a HTC approach which uses a PLM and a topic model to extract features from text documents which are used to train a classification model. Our objective is to determine whether the combination of the features extracted from the two models is beneficial to HTC performance in general. In our approach, the extracted features are passed through separate convolutional layers whose outputs are combined and passed to a label-wise attention mechanisms which obtains label-specific document representations by weighing the most important features for each class separately. We perform comprehensive experiments on three HTC benchmark datasets and show that using the features extracted from the topic model generally decreases classification performance compared to only using the features obtained by the PLM. In contrast to previous work, this shows that the incorporation of features extracted from topic models for text classification tasks should not be assumed beneficial.
\end{abstract}


\section{Introduction}
Hierarchical text classification (HTC) systems aim to assign text documents to a set of classes from a hierarchical class structure. Therefore, HTC systems can be used to improve the organisation and navigation of large document collections since they allow users to reduce their search scope from a large number of documents to a smaller subset of documents with finer-grained categories. These systems also allow users to select the level of granularity that they prefer based on the level of the class hierarchy.

Recently proposed HTC approaches attempt to combine the hierarchical class structure information with pre-trained language models (PLMs) through various techniques~\citep{Chen,Huang2,Jiang,dutoit2,Wang1,Wang2}. These approaches typically fine-tune the parameters of the PLMs to leverage the language understanding capabilities of the PLM while adjusting the parameters for the specific downstream task. However, PLMs can also be used as feature extraction models by passing a text token sequence through the PLM to obtain semantic and contextual representations for each token in the sequence without updating the parameters of the model. These features can be used to train a new model which aims to learn the functional mapping from the extracted features to the expected output.

Topic models~\citep{topicmodels} can also be used to extract features from a text document by creating abstract topics which are represented as a distribution of words in a corpus of documents. Topic models can be used to represent each document and word as a distribution over the set of abstract topics such that these representations capture the corpus-level topics.

\citet{liu2021talbert} proposed a text classification approach which combines the semantic representations extracted from a PLM with the topic representations extracted from a topic model. They used a PLM and a LDA topic model~\citep{lda} to extract features from text documents to train a convolutional neural network (CNN)~\citep{cnn} classifier. Their reasoning behind the combination of these feature extraction models is that the representations obtained from the PLM capture the granular semantic and contextual information of the document while the topic model extracts higher-level ``global'' information which considers the entire corpus of documents. Their results showed that by combining the features obtained from the two models they were able to improve classification performance on multi-label text classification tasks. They claim that combining the semantic- and contextually-aware token representations from the PLM with the ``global'' topic representations of a LDA topic model improves performance since the features capture different granularities and attributes which can be used to distinguish between the classes of the text documents.

The combination of document representations extracted from a PLM and topic model has not been investigated for HTC tasks in previous work. Therefore, in this paper, we evaluate the efficacy of using a PLM and a topic model to extract features from text documents which are used to train a classifier model for HTC tasks. Our aim of this investigation is to determine whether the improvements obtained by the combination of features which capture different granularities of the document generalises from the text classification approach proposed by \citet{liu2021talbert} to HTC tasks. Therefore, we evaluate whether topic models may provide valuable information for distinguishing between different classes as opposed to the standard approach of only using a PLM to extract the semantic representations of the documents.

Our experimental setup uses a BERT~\citep{Devlin} PLM and a BERTopic~\citep{grootendorst2022bertopic} model to extract features from a document which are separately passed through dedicated convolutional layers. Subsequently, the output of the convolutional layers are combined and passed through label-wise attention and classification layers to obtain the final output of the model. We evaluate this model on three benchmark HTC datasets and show that using the topic model features generally leads to a decrease in classification performance compared to only using the PLM features. Furthermore, the results show that using a PLM as a feature extractor to train a CNN with label-wise attention consistently performs worse than recently proposed approaches which fine-tune the parameters of the PLM. This shows that the incorporation of topic model features for text classification tasks should not be assumed beneficial and should be carefully considered and evaluated.

\section{Background}
\subsection{Hierarchical Text Classification}
HTC approaches define the class hierarchy as a directed acyclic graph $\mathcal{H} = (C, E)$. The set of class nodes is given as $C = \{c_1, \ldots, c_L\}$, where $L$ is the total number of classes. The set of edges that represent the parent-child relationships between the class nodes are given as $E$. In this paper, we only use class hierarchies where each non-root node has a single parent node which results in a class taxonomy with a tree structure. The objective of HTC approaches is to classify a text document which contains $T$ tokens $\mathbf{x} = [x_1, \ldots, x_T]$ into a class set $Y' \subseteq C$ which constitutes one or more paths in $\mathcal{H}$.

\subsection{BERT}
\citet{Devlin} proposed the Bidirectional Encoder Representations from Transformers (BERT) language model which comprises a stack of transformer encoder modules~\citep{Vaswani}. BERT uses two self-supervised pre-training tasks: masked language modelling (MLM) and next sentence prediction (NSP). The model uses the MLM pre-training task to randomly mask 15\% of the input tokens with the objective of predicting their true tokens during training in order to obtain a general language understanding. Therefore, 15\% of tokens are replaced with a special \texttt{[MASK]} token after which the sequence is passed to the model which obtains the final hidden state for each token as the output of the last encoder module. These final hidden states capture the bidirectional context and semantics of each token through the stacked transformer encoder modules. The final hidden states of the \texttt{[MASK]} tokens are passed through a fully-connected layer (FCL) with a softmax function to obtain the output as a distribution over the entire vocabulary of tokens and these outputs are compared to the true tokens to drive the training process. Furthermore, the NSP task is used to pre-train the model by predicting the relationship between two sentences, i.e., whether the two sentences follow each other or not. BERT uses these techniques to extract semantic information by using the surrounding text of a given token to establish context. Through comprehensive experiments~\citep{Devlin} showed that BERT outperformed previous state-of-the-art approaches on various NLP tasks after fine-tuning the model for those tasks.

\subsection{Topic Models}
Topic models~\citep{topicmodels} are unsupervised machine learning models that have the objective of extracting abstract topics from a corpus of documents. The extracted topics are typically represented as a distribution over the vocabulary found in the corpus such that this distribution indicates the most important words to distinguish between the different topics. These topics can be used to represent a document or a particular word in a document as a distribution over the set of topics such that the distribution represents the most relevant topics associated with the document or word. Several topic modelling approaches have been proposed which include older approaches such as Latent Dirichlet Allocation (LDA)~\citep{lda} and more recent approaches which use PLMs such as BERTopic~\citep{grootendorst2022bertopic}.

LDA~\citep{lda} is a topic model based on the bag of words approach which represents a text document as a multi-set of its words and ignores ordinal and contextual information. The LDA model represents each topic as a probabilistic distribution over a set of words contained in the corpus of documents. The documents are represented as a probabilistic distribution over a set of topics and therefore the most relevant topics associated with each document can be determined based on the assumption that documents with similar topics will use a similar group of words. The LDA algorithm is provided with a fixed number of topics to discover in a corpus of documents and has the objective of learning the topic distribution of each document and the word distribution of each topic. Initially, the LDA algorithm randomly assigns each word in a document to a topic. These assignments of words to topics are used to form the topic representations as the distribution of the words assigned to each topic. Similarly, the document representations are formed as the distribution of the topics assigned to the words of the document. The algorithm attempts to reassign topics to words such that the topic representations become representative of the topics found in the corpus of documents.

BERTopic~\citep{grootendorst2022bertopic} is a more recent topic extraction model which utilises PLMs and clustering techniques to extract abstract topics from a corpus of documents. Firstly, BERTopic uses a Sentence-BERT~\citep{Reimers2019SentenceBERTSE} model to convert each document into a fixed-sized vector representation which captures the semantics of the document. A sentence-BERT model is a modified BERT model that is tuned to produce a semantic embedding for a sentence or document as opposed to the standard BERT model that produces embeddings for each token in the input sequence. This is done by passing two text sequences separately through the same BERT model and applying a contrastive loss function such that semantically similar sequences are pulled closer while semantically dissimilar sequences are forced further apart in the output embedding space. The BERTopic model reduces the dimensionality of these document representations with the UMAP~\citep{McInnes2018UMAP} dimensionality reduction algorithm and clusters the resulting representations with the HDBSCAN~\citep{McInnes2017HDBSCAN} algorithm, such that each cluster of documents represents a topic. In a final step, BERTopic uses these clusters to find the topic-word distributions by concatenating all of the documents in a cluster and applying a cluster-based TF-IDF method to find the most relevant words for the associated clusters. BERTopic uses the TF-IDF scores of each word to represent the distribution of words for the topic.

\section{Related Research}
\subsection{Hierarchical Text Classification}
In this section we discuss the recently proposed HTC approaches which are most relevant to this paper. HTC approaches are typically grouped into three categories which include flat, local, and global approaches.

Flat classifier approaches do not consider the hierarchical class structure but rather convert the HTC task into a multi-label classification task by flattening the class hierarchy. \citet{Mullenbach} proposed the Convolutional Attention for Multi-Label classification (CAML) model which comprises a CNN followed by a label-wise attention mechanism. The label-wise attention mechanism uses the features extracted from the CNN to obtain label-specific document representations by placing more weight on the most relevant features for each class separately through a dot product attention function. Later, \citet{Vu} proposed the Label Attention Model (LAAT) which uses LSTMs~\citep{lstm} with label-wise attention mechanisms that use an adaption of the general attention function. \citet{Vu} also proposed the JointLAAT model which splits the attention mechanisms for the different levels of the class hierarchy and converts the output predictions at a particular level to a vector which is concatenated to the lower-level label-specific document representations. \citet{LiuHiLAT} proposed the Hierarchical Label-wise Attention Transformer (HiLAT) model which separates a document into chunks and passes these chunks through a pre-trained XLNet model~\citep{XLNet} after which it sequentially applies token- and chunk-level label-wise attention mechanisms with the general attention function proposed by \citet{Vu}.

Local classifier approaches have the objective of using the structured class hierarchy information to improve performance by training multiple models and combining their predictions through different techniques~\citep{Banerjee,KollerSahami,Kowsari,Shimura,dutoit2024introducing}. For example, \citet{Kowsari} developed the Hierarchical Deep Learning for Text Classification (HDLTex) model which creates a deep neural network model for the non-leaf nodes of the class hierarchy. Furthermore, local approaches have been proposed which transfer the parameters of the classifiers in higher-levels to the lower-level classifiers since the higher-level classifiers form a good starting point for the lower-level classifiers which are then fine-tuned with the training data for their particular level~\citep{Banerjee,Shimura}.

Global approaches attempt to use a single model which captures the hierarchical class structure information to improve classification performance on HTC tasks. Global approaches have been proposed which use various techniques such as recursive regularisation~\citep{GopalYang}, reinforcement learning~\citep{Mao}, meta-learning~\citep{Wu}, and capsule networks~\citep{Peng}. Recent approaches have shown that encoding the class hierarchy with a graph neural network and incorporating this representation into the classification process can improve performance~\citep{Chen,Deng,Jiang,dutoit2, Wang1,Wang2,Zhou}. For example, \citet{Chen} proposed the hierarchy-aware label semantics matching network (HiMatch) which converts the HTC task into a semantic matching task by projecting the text document and the class hierarchy into the same embedding space. Alternatively, \citet{Wang1} proposed the Hierarchy Guided Contrastive Learning (HGCLR) model which  incorporates the structured class hierarchy information into a PLM through a contrastive learning objective function. More recently, \citet{Huang2} introduced the Hierarchy-Aware T5 model with Path-Adaptive Attention Mechanism (PAAMHiA-T5) which generates labels through a T5 model~\citep{Raffel} and uses a path-adaptive attention mechanism to allow the classifier to focus on the hierarchical path of the current class. \citet{Jiang} proposed Hierarchy-guided BERT with Global and Local hierarchies (HBGL) which extends the input text token sequence that is passed to BERT by concatenating the sequence with hierarchical class embeddings and a mask token for each level of the class hierarchy such that the model attempts to predict the classes at each level through the mask tokens. Similarly, \citet{Wang2} proposed the Hierarchy-aware Prompt Tuning (HPT) approach which converts the input token sequence through the addition of prompts for each level of the class hierarchy followed by a masked token such that the HTC task is converted to the MLM task used during pre-training. Recently, \citet{duToitLabelwise} proposed an approach which fine-tunes a PLM through the use of a hierarchical label-wise attention mechanism which leverages the predictions of all higher levels during the prediction of classes at a particular level.  More recently, \citet{dutoit2} also proposed the Hierarchy-aware Prompt Tuning for Discriminative PLMs (HPTD) approach which which extends the HPT approach to discriminative language models such as ELECTRA~\citep{electra} and DeBERTaV3~\citep{He2021DeBERTaV3}.

\subsection{Combining Language and Topic Models}
\citet{peinelt2020tbert} proposed the topic-informed BERT-based model (tBERT) which combines information extracted from a LDA topic model with a BERT PLM for semantic similarity prediction tasks which have the objective of measuring the semantic similarity between text documents. To determine the semantic similarity between two documents, tBERT first concatenates the two documents and passes them through a BERT model to obtain the sentence pair representation from the final hidden state of the \texttt{[CLS]} token. Furthermore, tBERT uses a topic model to extract abstract topics from the corpus of documents and represents each document as a distribution over these topics such that they capture the most important topics associated with each document. They concatenate the representations from the BERT and topic models and pass these features through a FCL to obtain the output of the model which represents the semantic similarity of the documents. They show that combining the representations extracted from a topic model and a PLM improved performance over multiple semantic similarity prediction tasks, especially in domain-specific cases.

\citet{liu2021talbert} proposed an approach which combines the features extracted from a BERT-based PLM and a LDA topic model to train a CNN classifier for multi-label text classification tasks. The rationale for combining the features obtained by these models is that the PLM extracts fine-grained semantic and contextual representations while the representations obtained by the topic model capture higher-level ``global'' information which takes the entire corpus of documents into account. The PLM is used to extract semantic word embeddings for each token in the document and the document as a whole. Furthermore, the topic model extracts vector representations for each word and document which represents the distribution over the set of abstract topics created by the topic model. They combine the extracted features from these models to train a CNN classifier which comprises multiple convolutional layers with different filter sizes to group varying lengths of token sequences together. Finally, their approach applies max-pooling to the output of the convolutional layers followed by a FCL to obtain the confidence scores of a document belonging to each of the possible classes.

\section{Methodology}
\begin{figure}[h]
    \centerline{\scalebox{0.92}{\input{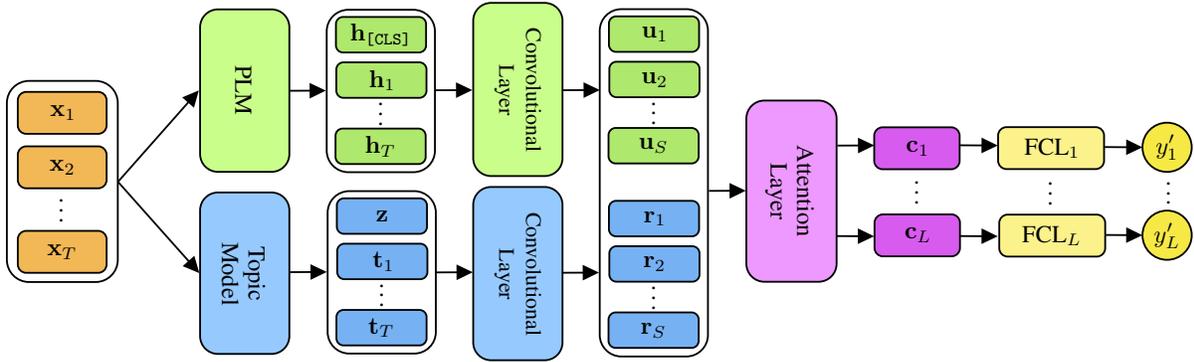}}}
    \caption{Model architecture. The input text token sequence (orange) is separately passed through the PLM and topic model to obtain local semantic embeddings (green) and global topic embeddings (blue) for each token in the input sequence and the sequence as a whole. These features are passed through separate convolutional layers whose outputs are combined and passed through a label-wise attention layer which obtains label-specific representations of the document (purple). Finally, these representations are passed through their associated label-wise FCLs which calculate the confidence of the document belonging to each of the possible classes (yellow).}
    \label{fig:model_architecture}
\end{figure}

Figure~\ref{fig:model_architecture} illustrates the high-level architecture of our HTC methodology which combines representations from a PLM and a topic model to train a model which comprises convolutional layers followed by label-wise attention and classification layers.

\subsection{Feature Extraction}
Suppose we have a text token sequence $\mathbf{x} = [x_1, \ldots, x_T]$, where $T$ is the number of tokens in the sequence. We extract local semantic and global topic features for each token and the sequence as a whole through a PLM and topic model respectively. 

We extract local semantic representations by passing the text token sequence through a PLM to obtain the final hidden states of the token sequence as $\mathbf{H} = [\mathbf{h}_1, \ldots, \mathbf{h}_T] \in \mathbb{R}^{T \times d_h}$, where $\mathbf{h}_t \forall t \in \{1, \ldots, T\}$ is the final hidden state for token $x_t$ and $d_h$ is the dimension of the hidden states. These embeddings provide ``local'' context and semantics of the tokens in a document without considering the other documents in the corpus. Furthermore, the PLM extracts the document-level information through the final hidden state of the \texttt{[CLS]} token as $\mathbf{h}_\texttt{[CLS]} \in \mathbb{R}^{d_h}$.

The topic model creates $d_f$ abstract topics where each topic is a distribution over the words in the corpus of documents such that it captures the semantics of the topic. We obtain a document-level topic vector $\mathbf{z} \in \mathbb{R}^{d_f}$ which is a distribution over the $d_f$ topics obtained by the topic model such that $\mathbf{z}$ represents the information of the document in terms of the abstract topics. Similarly, each word in the document is represented as a distribution over the set of abstract topics such that topic representations are obtained for the document words as $\mathbf{T} = [\mathbf{t}_1, \ldots, \mathbf{t}_T] \in \mathbb{R}^{T \times d_f}$, where $\mathbf{t}_t \forall t \in \{1, \ldots, T\}$ is the topic vector for token $x_t$. Each of these vectors represent the confidence of the token or document belonging to each of the topics created by the topic model. Therefore, the topic representations provide a ``global'' topic representation of each word and document by considering the context of the topics found in the corpus as a whole.

\subsection{Convolutional Layer}
The feature vectors obtained by the language and topic models are passed through separate convolutional layers with identical architectures to extract patterns from the respective sequences.

We use a convolutional layer which comprises $P$ convolutional channels with different filter sizes, where each channel groups together different length sequences of word and document embeddings to extract various patterns from the feature representations of the document. Furthermore, we use $Q$ convolutional filters for each convolutional channel, such that each filter learns different patterns from the input sequence to produce $Q$ feature vectors. We add padding (zero vectors) to the start and end of the token sequence, such that all of the convolutional filters step over the token sequence to obtain a $T$-dimensional output vector $\mathbf{q}_i \forall i \in \{1, \ldots, Q\}$. Each convolutional channel combines the outputs from its convolutional filters to obtain a feature matrix $\mathbf{Q} = [\mathbf{q}_1, \ldots, \mathbf{q}_Q] \in \mathbb{R}^{Q \times T}$. We concatenate the features obtained by each convolutional channel to form the output of the convolutional layer.

We prepend the final hidden state of the \texttt{[CLS]} token to the final hidden states of the token sequence ($\mathbf{H}$) and pass it to the convolutional layer associated with the PLM to obtain the output as $\mathbf{U} = [\mathbf{u}_1, \ldots, \mathbf{u}_S] \in \mathbb{R}^{S \times T}$ where $S = P \cdot Q$. Similarly, we prepend the document topic vector ($\mathbf{z}$) to the sequence of token topic vectors ($\mathbf{T}$) and pass it through its associated convolutional layer to obtain the output as $\mathbf{R} = [\mathbf{r}_1, \ldots, \mathbf{r}_S] \in \mathbb{R}^{S \times T}$. Finally, we concatenate the output from the two separate convolutional layers to form $\mathbf{G} = [\mathbf{u}_1, \ldots, \mathbf{u}_S, \mathbf{r}_1, \ldots, \mathbf{r}_S] \in \mathbb{R}^{2S \times T}$.

\subsection{Attention Layer}

For our standard architecture shown in Figure~\ref{fig:model_architecture} we consider the label-wise attention mechanism from \citet{Vu} (LAAT) to convert the concatenated feature vectors from the convolutional layers ($\mathbf{G}$) to a matrix $\mathbf{C} \in \mathbb{R}^{L \times T}$, where $L$ is the number of classes and each row $\mathbf{c}_j \forall j \in \{1, \ldots, L\}$ is a label-specific vector representation of the text document. We refer to this attention mechanism as the general attention (GA) mechanism which calculates an attention weight matrix $\boldsymbol{\alpha} \in \mathbb{R}^{L \times 2S}$ to represent the attention weights of each feature for each class as follows:
\begin{align}
    \mathbf{Z} = \text{tanh}(\mathbf{Q_{\text{GA}}}\mathbf{G}^T) \\
    \boldsymbol{\alpha} = \text{softmax}(\mathbf{U}_{\text{GA}}\mathbf{Z})
\end{align}
\noindent where $\mathbf{Q}_{\text{GA}} \in \mathbb{R}^{d_p \times T}$ and $\mathbf{U_{\text{GA}}} \in \mathbb{R}^{L \times d_p}$ are weight matrices that are learnt during training and $d_p$ is a chosen hyperparameter.

The attention weight matrix is used to calculate the label-wise representations of the document as follows:

\begin{equation}
    \mathbf{C} = \boldsymbol{\alpha}\mathbf{G}
\end{equation}

\noindent where the $j$-th row ($\mathbf{c}_j$) of the matrix $\mathbf{C} \in \mathbb{R}^{L \times T}$ represents the information of the document with regard to class $j$.  

\subsection{Output Layer}
Following previous label-wise attention approaches~\citep{Mullenbach,Vu,StrydomPaper,duToitLabelwise}, each row of the $\mathbf{C}$ matrix is passed through a separate FCL with a sigmoid activation function ($\sigma$) such that the confidence score for class $j$ is calculated as: 
\begin{equation}
    y_j' = \sigma(\mathbf{w}_j \cdot \mathbf{c}_{j} + b_j) 
\end{equation}

\noindent where $\mathbf{c}_j$ is the $j$-th row in $\mathbf{C}$. The weight vectors $\mathbf{w}_1, \ldots, \mathbf{w}_L$ and bias units $b_1, \ldots, b_L$ are combined into a weight matrix $\mathbf{W} \in \mathbb{R}^{L \times T}$ and bias vector $\mathbf{b}_W \in \mathbb{R}^{L}$ such that the output of the model is calculated as:
\begin{equation}
    \mathbf{y}' = \sigma\bigl((\mathbf{W} \otimes \mathbf{C})\mathbf{1} + \mathbf{b}_W\bigr)
\end{equation}

\noindent where $\mathbf{1} \in \mathbb{R}^{T}$ is a vector of ones and $\otimes$ is the element-wise multiplication operator.

The objective during training is to minimize the binary cross-entropy loss function which is calculated as:
\begin{equation}
\mathcal{L} = - \sum_{j=1}^L \bigl(y_j \text{log} (y'_j) + (1-y_j) \text{log} (1-y'_j)\bigr)
\end{equation}
\noindent where the ground truth label for class $j$ is given as $y_j \in \{0, 1\}$.

Finally, the output scores can be used to obtain the multi-hot vector $\mathbf{Y}' = [Y'_{1}, \ldots, Y'_{L}]$ which represents the classes that a document is assigned to and is calculated as:
\begin{equation}
    Y'_{j} = \begin{cases}
        1, & y'_j \geq \gamma \\
        0, & y'_j < \gamma
    \end{cases}
\end{equation}
\noindent where $\gamma$ is a chosen threshold.

\subsection{Hierarchical Attention Layer}
We also use the global hierarchical label-wise attention (GHLA) mechanism proposed by \citet{duToitLabelwise} which splits the attention mechanisms for the different levels of the class hierarchy and uses all of the higher-level prediction information during the prediction task at a particular level.

Suppose we have a particular HTC task which has a hierarchical class structure with $K$ levels, and $N_k$ is the number of classes in level $k \in \{1, \ldots, K\}$. GHLA obtains label-specific document representations for each first-level class $j \in \{1, \ldots, N_1\}$ by passing the concatenated feature vectors from the convolutional layers ($\mathbf{G}$) through the first-level attention layer (Attention$_1$) to obtain $\mathbf{C}_{1} = [\mathbf{c}_{1, 1}, \ldots, \mathbf{c}_{1, N_1}]$ where $\mathbf{c}_{1, j}$ is a vector that represents the document for the $j$-th class in level one. GHLA passes the first-level class representations through associated FCLs to obtain the output confidence scores for level one as $\mathbf{y}'_1 = [y'_{1, 1}, \ldots, y'_{1, N_1}]$, where $y'_{1, j}$ is the confidence score of the $j$-th class in level one. These predictions are then passed through another FCL to obtain a vector representation of level one as $\mathbf{s}_1$. GHLA concatenates this first level prediction representation ($\mathbf{s}_1$) to the second-level label-wise document representations ($\mathbf{C}_2$) obtained by the associated attention layer (Attention$_2$) as follows:
\begin{equation}
    \mathbf{D}_2 = [Concat(\mathbf{c}_{2, 1}, \mathbf{s}_1), \ldots, Concat(\mathbf{c}_{2, N_2}, \mathbf{s}_1)]
\end{equation}
\noindent where $Concat$ concatenates the vectors. Subsequently, the rows of $\mathbf{D}_2$ are passed through separate FCLs to obtain the predictions for the second-level classes which are passed through another FCL to obtain the the second-level prediction representation $\mathbf{s}_2$. GHLA concatenates all of the ancestor-level prediction representations to the label-wise document representations at a particular level such that the third-level document representations are formed as follows:
\begin{equation}
    \mathbf{D}_3 = [Concat(\mathbf{c}_{3, 1}, \mathbf{s}_1, \mathbf{s}_2), \ldots, Concat(\mathbf{c}_{3, N_3}, \mathbf{s}_1, \mathbf{s}_2)]
\end{equation}
This procedure is continued until the final level is reached and this allows the prediction task at a particular level to use the prediction representations of all the higher levels such that the hierarchical class structure information is leveraged during the classification process.

\section{Experiments}
\subsection{Datasets and Evaluation Metrics}

\begin{table}[t!]
    \centering
    \caption{Characteristics of the benchmark datasets. The columns ``Levels'' and ``Classes'' give the number of levels and classes in the class structure. ``Avg.~Classes'' is the average number of classes per document, while ``Train'', ``Dev'',  and ``Test'' are the number of instances in each of the dataset splits.}
    \label{tab:Datasets}
    \begin{tabular}{|c|r|r|r|r|r|r|}
        \hline
        \normalsize Dataset & \normalsize Levels & \normalsize Classes $d_f$ & \normalsize Avg.~Classes & \normalsize Train & \normalsize Dev & \normalsize Test \\
        \hline
        WOS & 2 & 141 & 2.00 & 30,070 & 7,518 & 9,397 \\
        RCV1-V2 & 4 & 103 & 3.24 & 20,833 & 2,316 & 781,265 \\
        NYT & 8& 166 & 7.60 & 23,345 & 5,834 & 7,292 \\
        \hline
    \end{tabular}
\end{table}

We evaluate all models with the most commonly used benchmark HTC datasets which are: the Web of Science (WOS)~\citep{Kowsari}, the RCV1-V2 \citep{RCV1}, and the NYTimes (NYT)~\citep{nyt} datasets. The WOS dataset comprises abstracts of research publications from the Web of Science publication database, whereas RCV1-V2 and NYT contain news articles from Reuters and The New York Times respectively. To facilitate fair comparisons, we follow the same dataset splits and processing steps as in previous work~\citep{Zhou,Deng,Chen,Jiang,Huang2,Wang1,Wang2}. Tables~\ref{tab:Datasets} and~\ref{tab:BranchingFactors} provide the summary statistics and hierarchical properties of these datasets. These tables show that the WOS dataset has the simplest class hierarchy with only two levels, while the NYT dataset has the most complex class hierarchy with eight levels and 166 classes. Furthermore, the three datasets have a similar number of training instances, however, the RCV1-V2 dataset has a significantly larger testing set.

\begin{table}[ht]
    \caption{The average per-level branching factor of the hierarchy in each benchmark dataset, which is calculated as the average number of child nodes for the nodes at a particular level. The node counts per level are given in parentheses.}
    \label{tab:BranchingFactors}
    \centering
    \begin{tabular}{|l|r|r|r|r|r|r|r|r|}
        \hline
        \normalsize Dataset & \multicolumn{1}{l|}{\normalsize Level 1} & \multicolumn{1}{l|}{\normalsize Level 2} & \multicolumn{1}{l|}{\normalsize Level 3} & \multicolumn{1}{l|}{\normalsize Level 4} & \multicolumn{1}{l|}{\normalsize Level 5} & \multicolumn{1}{l|}{\normalsize Level 6} & \multicolumn{1}{l|}{\normalsize Level 7} & \multicolumn{1}{l|}{\normalsize Level 8} \\
        \hline
        WOS & 19.14 (7) & 0.00 (134) & \text{--} & \text{--} & \text{--} & \text{--} & \text{--} & \text{--}\\
        RCV1-V2 & 13.75 (4) & 0.78~~(55) & 0.02 (43) & 0.00~~(1) & \text{--} & \text{--} & \text{--} & \text{--} \\
        NYT & 6.75 (4) & 1.89~~(27) & 0.92 (51) & 0.36 (47) & 0.71 (17) & 0.50 (12) & 0.33 (6) & 0.00 (2) \\
    \hline
    \end{tabular}
\end{table}

In all of our experiments we use the standard Micro-F1 and Macro-F1 evaluation metrics since these are the most common metrics used for HTC tasks~\citep{Chen,Deng,Huang2,Jiang,Wang1,Wang2,Zhou}. Micro-F1 averages the performance over all instances while Macro-F1 equally weights the performance for each class.

\subsection{Implementation Details}
We used the \texttt{bert-base-uncased} model from Hugging Face as the PLM in all of our experiments which is the standard BERT model used in previous approaches. We used the Adam optimiser~\citep{kingma2017adam} and performed hyperparameter tuning on the learning rate and batch size with possible values of $\{1\text{e-}4, 5\text{e-}4, 1\text{e-}3\}$ and $\{16, 32\}$ respectively. Furthermore, we chose $d_p = 256$, $\gamma = 0.5$, and $d_f$ as the number of classes in the associated dataset as shown in Table~\ref{tab:Datasets}.


We refer to the two models under evaluation as Topic Attention CNN (TopAttCNN) such that \textbf{TopAttCNN$_\text{GA}$} and \textbf{TopAttCNN$_\text{GHLA}$} are the architectures which use the GA and GHLA mechanisms respectively. We evaluated these two models using a GA layer for each level of the GHLA mechanism. Furthermore, as a baseline, we evaluated a model that does not incorporate the topic model information, i.e., it only uses the embeddings from the PLM with its associated convolutional layer, to observe the impact that the topic embeddings have on performance. We investigated both of the attention mechanisms for this setting and refer to these models as \textbf{AttCNN$_\text{GA}$} and \textbf{AttCNN$_\text{GHLA}$}.

We trained the models on the training set for a maximum of 20 epochs and stopped training when the harmonic mean of the Micro-F1 and Macro-F1 on the development set did not increase for 5 consecutive epochs. We chose the models and hyperparameter combination which obtained the highest harmonic mean of Micro-F1 and Macro-F1 on the development set to evaluate the performance of our model on the test set. Our results are reported as the average scores over three runs with different random seeds.

\subsection{Main Results}
Table~\ref{tab:MainResults} presents the performance results of the models under evaluation compared to the recent HTC approaches. The results show that the addition of the topic model information generally decreases the performance of the model. The AttCNN$_\text{GHLA}$ model achieves the highest Macro-F1 score on each of the datasets along with the highest Micro-F1 on WOS (85.00) and comparable performance to the best-performing models for Micro-F1 on RCV1-V2 (84.54) and NYT (76.94). AttCNN$_\text{GHLA}$ only uses the extracted features from the PLM to classify the text documents, showing that the addition of global topic information does not improve classification performance in general. We hypothesise that the topic information of the tokens in the text sequence are often not useful for distinguishing between the different classes since the extracted topics may not correlate with the classes for the particular dataset.

The results show that the GHLA mechanism generally outperforms GA when removing the topic information (AttCNN) whereas the TopAttCNN$_\text{GA}$ model outperforms TopAttCNN$_\text{GHLA}$ on all performance measures apart from the Micro-F1 score on NYT. Furthermore, we show that the difference in performance between our four models is very small, with most of the metrics having less than a 1\% difference. The only metric with a notably large difference is the Macro-F1 score for the TopAttCNN$_\text{GHLA}$ model on the RCV1-V2 dataset, which is significantly lower (2.79) than the best-performing model. Finally, we observe that the recently proposed HTC approaches outperform the combination models under evaluation on each of the datasets, with large margins for Macro-F1 on the RCV1-V2 and NYT datasets. 

\begin{table}[ht]
    \caption{Performance comparisons of the AttCNN and TopAttCNN models using the three commonly used benchmark datasets.}
    \centering
    \label{tab:MainResults}
    \begin{tabular}{|l|c|c|c|c|c|c|}
        \hline
         &
        \multicolumn{2}{|c|}{\normalsize WOS} &
        \multicolumn{2}{c}{\normalsize RCV1-V2} &
        \multicolumn{2}{|c|}{\normalsize NYT} \\
        \normalsize Model & {Micro-F1} & {Macro-F1} & {Micro-F1} & {Macro-F1} & {Micro-F1} & {Macro-F1} \\
        \hline
        HiMatch \citep{Chen} &  86.20 & 80.53 & 84.73 & 64.11 & \text{--} & \text{--} \\
        HGCLR \citep{Wang1} & 87.11 & 81.20 & 86.49 & 68.31 & 78.86 & 67.96 \\
        PAAMHiA-T5\footnote{Results obtained using twice the number of model parameters as the other approaches.} \citep{Huang2} & \bf{90.36} & 81.64 & 87.22 & 70.02 & 77.52 & 65.97 \\
        HBGL \citep{Jiang} & \bf{87.36} & \bf{82.00} & 87.23 & \bf{71.07} & 80.47 & 70.19  \\
        HPT \citep{Wang2}  & 87.16 & 81.93 & 87.26 & 69.53 & 80.42 & 70.42 \\
        GHLA$_\text{RoBERTa}$ \citep{duToitLabelwise} & 87.00 & 81.44 & \bf{87.78} & 70.21 & \bf{81.41} & \bf{72.27} \\
        \hline
        AttCNN$_\text{GA}$ & 84.93 & 78.57  & 84.67 & 62.48 & 77.07  & 64.08 \\
        TopAttCNN$_\text{GA}$ & 84.76 & 78.07  & \bf{84.72} & 62.33  & 76.88  & 64.18 \\
        AttCNN$_\text{GHLA}$ & \bf{85.00}  & \bf{79.02} & 84.54 & \bf{63.11} & 76.94  & \bf{64.57} \\
        TopAttCNN$_\text{GHLA}$ & 84.64 & 77.86 & 84.51 & 60.32  & \bf{77.08} & 64.35 \\
        \hline
    \end{tabular}
\end{table}

\subsection{Stability Analysis}
Table~\ref{tab:sensitivity_analysis} presents the mean scores and standard deviations over three runs with different random seeds for each model. The results show no clear best-performing model in terms of stability over multiple runs, with most standard deviations being very similar across the three datasets. However, the TopAttCNN$_\text{GHLA}$ obtained a significantly higher standard deviation on the RCV1-V2 Macro-F1 (1.34), which shows that this methodology may produce inconsistent results for datasets with a high class imbalance such as RCV1-V2.

\begin{table}[h]
    \caption[The average performance results of evaluating the AttCNN and TopAttCNN models over three independent runs.]{The average performance results of evaluating the AttCNN and TopAttCNN models over three independent runs. The values in parentheses show the corresponding standard deviations.}
    \label{tab:sensitivity_analysis}
    \centering
    \begin{tabular}{|l|c|c|c|c|c|c|}
        \hline
        &
        \multicolumn{2}{|c}{\normalsize WOS} &
        \multicolumn{2}{|c}{\normalsize RCV1-V2} &
        \multicolumn{2}{|c|}{\normalsize NYT} \\
        Model & {Micro-F1} & {Macro-F1} & {Micro-F1} & {Macro-F1} & {Micro-F1} & {Macro-F1} \\
        \hline
        AttCNN$_\text{GA}$ & 84.93 (0.23) & 78.57 (\textbf{0.25}) & 84.67 (0.11) & 62.48 (0.29) & 77.07 (0.28) & 64.08 (0.35)\\
        TopAttCNN$_\text{GA}$ & 84.76 (0.18) & 78.07 (0.46) & 84.72 (\textbf{0.10}) & 62.33 (0.28) & 76.88 (\textbf{0.22}) & 64.18 (0.36) \\
        AttCNN$_\text{GHLA}$ & 85.00 (0.27) & 79.02 (0.34) & 84.54 (0.13) & 63.11 (\textbf{0.05}) & 76.94 (0.42) & 64.57 (0.49) \\
        TopAttCNN$_\text{GHLA}$ & 84.64 (\textbf{0.17}) & 77.86 (0.37) & 84.51 (0.11) & 60.32 (1.34) & 77.08 (0.33) & 64.35 (\textbf{0.25}) \\
        \hline
    \end{tabular}
    
\end{table}

\subsection{Level-wise Results}
\begin{figure}[ht!]
    \begin{subfigure}{0.48\textwidth}
    \centerline{\begin{tikzpicture}[scale = 0.58, font = \Large]
  \begin{axis}[
    symbolic x coords ={Level 1,Level 2},
    axis y line*=left,
    ybar,
    ylabel={Micro-F1},
    ymin = 75, ymax = 100,
    xlabel near ticks,
    xtick = data,
    ylabel near ticks,
    bar width=0.8cm,
    enlarge x limits= 0.5,
    nodes near coords align={center},
    width = 12cm,
    height=8.5cm,
  ]
\addplot[draw = black, fill = cyan] 
    coordinates {(Level 1, 89.69) (Level 2, 79.95)};
\addplot[draw = black, fill = lime]  
    coordinates {(Level 1, 89.49) (Level 2, 80.31)};
\addplot[draw = black, fill = orange]  
    coordinates {(Level 1, 89.41) (Level 2, 79.89)};
\addplot[draw = black, fill = pink] 
    coordinates {(Level 1, 89.23) (Level 2, 79.82)};

\legend {AttCNN$_\text{GA}$, AttCNN$_\text{GHLA}$, TopicAttCNN$_\text{GA}$, TopicAttCNN$_\text{GHLA}$};
\end{axis}
  
\begin{axis}[
    symbolic x coords ={Level 1,Level 2},
    ymin = 0, ymax = 5000,
    hide x axis,
    axis y line*=right,
    yticklabel pos=right,
    ylabel={Average training instances per class},
    ylabel near ticks,
    nodes near coords align={center},
    enlarge x limits=0.5,
    width = 12cm,
    height=8.5cm,
  ]
    \addplot[color = black, mark = square*] coordinates {(Level 1, 4295) (Level 2, 224)};
\end{axis}
  
\end{tikzpicture}}
    \caption{WOS Micro-F1.}
    \label{fig:LevelWiseWOSMicro}
    \end{subfigure}
    \hfill
    \begin{subfigure}{0.48\textwidth}
    \centerline{\begin{tikzpicture}[scale = 0.58, font = \Large]
  \begin{axis}[
    symbolic x coords ={Level 1,Level 2},
    axis y line*=left,
    ybar,
    ylabel={Macro-F1},
    ymin = 75, ymax = 100,
    xlabel near ticks,
    xtick = data,
    ylabel near ticks,
    bar width=0.8cm,
    enlarge x limits= 0.5,
    nodes near coords align={center},
    width = 12cm,
    height=8.5cm,
  ]
\addplot[draw = black, fill = cyan] 
    coordinates {(Level 1, 89.80) (Level 2, 77.98)};
\addplot[draw = black, fill = lime]  
    coordinates {(Level 1, 89.60) (Level 2, 78.47)};
\addplot[draw = black, fill = orange]  
    coordinates {(Level 1, 89.59) (Level 2, 77.47)};
\addplot[draw = black, fill = pink] 
    coordinates {(Level 1, 89.42) (Level 2, 77.26)};

\legend {AttCNN$_\text{GA}$, AttCNN$_\text{GHLA}$, TopicAttCNN$_\text{GA}$, TopicAttCNN$_\text{GHLA}$};
\end{axis}
  
\begin{axis}[
    symbolic x coords ={Level 1,Level 2},
    ymin = 0, ymax = 5000,
    hide x axis,
    axis y line*=right,
    yticklabel pos=right,
    ylabel={Average training instances per class},
    ylabel near ticks,
    nodes near coords align={center},
    enlarge x limits=0.5,
    width = 12cm,
    height=8.5cm,
  ]
    \addplot[color = black, mark = square*] coordinates {(Level 1, 4295) (Level 2, 224)};
\end{axis}
  
\end{tikzpicture}}
    \caption{WOS Macro-F1.}
    \label{fig:LevelWiseWOSMacro}
    \end{subfigure}
    
    \vspace{1em}
    
    \begin{subfigure}{0.48\textwidth}
    \centerline{\begin{tikzpicture}[scale = 0.58, font = \Large]
  \begin{axis}[
    symbolic x coords = {Level 1,Level 2,Level 3,Level 4},
    axis y line*=left,
    ybar,
    ylabel={Micro-F1},
    ymin = 52, ymax = 100,
    xlabel near ticks,
    xtick = data,
    ylabel near ticks,
    bar width=0.4cm,
    enlarge x limits= 0.2,
    nodes near coords align={center},
    width=12cm,
    height=8.6cm,
  ]
\addplot[draw = black, fill = cyan] 
    coordinates {(Level 1, 92.71) (Level 2, 78.94) (Level 3, 82.50) (Level 4, 58.11)};
\addplot[draw = black, fill = lime]  
    coordinates {(Level 1, 92.57) (Level 2, 78.84) (Level 3, 82.12) (Level 4, 67.22)};
\addplot[draw = black, fill = orange]  
    coordinates {(Level 1, 92.74) (Level 2, 79.02) (Level 3, 82.49) (Level 4, 64.34)};
\addplot[draw = black, fill = pink] 
    coordinates {(Level 1, 92.56) (Level 2, 78.88) (Level 3, 81.78) (Level 4, 70.02)};

\legend {AttCNN$_\text{GA}$, AttCNN$_\text{GHLA}$, TopicAttCNN$_\text{GA}$, TopicAttCNN$_\text{GHLA}$};
\end{axis}
  
  \begin{axis}[
    symbolic x coords = {Level 1,Level 2, Level 3, Level 4},
    ymin = 0, ymax = 8000,
    hide x axis,
    axis y line*=right,
    yticklabel pos=right,
    ylabel={Average training instances per class},
    ylabel near ticks,
    nodes near coords align={center},
    enlarge x limits=0.2,
    width=12cm,
    height=8.6cm,
  ]
    \addplot[color = black, mark = square*] coordinates {(Level 1, 6096) (Level 2, 521) (Level 3, 300) (Level 4, 359)};
  \end{axis}
\end{tikzpicture}}
    \caption{RCV1-V2 Micro-F1.}
    \label{fig:LevelWiseRCV1Micro}
    \end{subfigure}
    \hfill
    \begin{subfigure}{0.48\textwidth}
    \centering
    \centerline{\begin{tikzpicture}[scale = 0.58, font = \Large]
  \begin{axis}[
    symbolic x coords = {Level 1,Level 2,Level 3,Level 4},
    axis y line*=left,
    ybar,
    ylabel={Macro-F1},
    ymin = 52, ymax = 100,
    xlabel near ticks,
    xtick = data,
    ylabel near ticks,
    bar width=0.4cm,
    enlarge x limits= 0.2,
    nodes near coords align={center},
    width=12cm,
    height=8.6cm,
  ]
\addplot[draw = black, fill = cyan] 
    coordinates {(Level 1, 91.21) (Level 2, 63.91) (Level 3, 58.07) (Level 4, 58.11)};
\addplot[draw = black, fill = lime]  
    coordinates {(Level 1, 91.13) (Level 2, 64.54) (Level 3, 58.59) (Level 4, 67.22)};
\addplot[draw = black, fill = orange]  
    coordinates {(Level 1, 91.28) (Level 2, 63.75) (Level 3, 57.78) (Level 4, 64.34)};
\addplot[draw = black, fill = pink] 
    coordinates {(Level 1, 91.12) (Level 2, 63.08) (Level 3, 53.71) (Level 4, 70.02)};

\legend {AttCNN$_\text{GA}$, AttCNN$_\text{GHLA}$, TopicAttCNN$_\text{GA}$, TopicAttCNN$_\text{GHLA}$};
\end{axis}
  
  \begin{axis}[
    symbolic x coords = {Level 1,Level 2, Level 3, Level 4},
    ymin = 0, ymax = 8000,
    hide x axis,
    axis y line*=right,
    yticklabel pos=right,
    ylabel={Average training instances per class},
    ylabel near ticks,
    nodes near coords align={center},
    enlarge x limits=0.2,
    width=12cm,
    height=8.6cm,
  ]
    \addplot[color = black, mark = square*] coordinates {(Level 1, 6096) (Level 2, 521) (Level 3, 300) (Level 4, 359)};
  \end{axis}
  
\end{tikzpicture}}
    \caption{RCV1-V2 Macro-F1.}
    \label{fig:LevelWiseRCV1Macro}
    \end{subfigure}

    \vspace{1em}

    \begin{subfigure}{0.48\textwidth}
    \centerline{\begin{tikzpicture}[scale = 0.58, font = \Large]
  \begin{axis}[
    symbolic x coords ={Level 1,Level 2,Level 3,Level 4,Level 5,Level 6,Level 7,Level 8},
    xticklabel style={font=\small},
    axis y line*=left,
    ybar,
    ylabel={Micro-F1},
    ymin = 40, ymax = 100,
    xlabel near ticks,
    xtick = data,
    ylabel near ticks,
    bar width=0.15cm,
    enlarge x limits=0.1,
    nodes near coords align={center},
    width=12cm,
    height=8.5cm,
  ]
\addplot[draw = black, fill = cyan] 
    coordinates {(Level 1, 87.92) (Level 2, 79.01) (Level 3, 71.82) (Level 4, 69.41) (Level 5, 72.19) (Level 6, 70.49) (Level 7, 69.36) (Level 8, 67.32)};
\addplot[draw = black, fill = lime]  
    coordinates {(Level 1, 87.73) (Level 2, 78.87) (Level 3, 71.80) (Level 4, 69.45) (Level 5, 72.34) (Level 6, 71.11) (Level 7, 68.84) (Level 8, 65.58)};
\addplot[draw = black, fill = orange]  
    coordinates {(Level 1, 87.77) (Level 2, 78.87) (Level 3, 71.39) (Level 4, 69.36) (Level 5, 72.00) (Level 6, 70.38) (Level 7, 68.60) (Level 8, 64.51)};
\addplot[draw = black, fill = pink] 
    coordinates {(Level 1, 87.86) (Level 2, 78.99) (Level 3, 71.73) (Level 4, 69.76) (Level 5, 72.26) (Level 6, 70.96) (Level 7, 69.04) (Level 8, 65.87)};

\legend {AttCNN$_\text{GA}$, AttCNN$_\text{GHLA}$, TopicAttCNN$_\text{GA}$, TopicAttCNN$_\text{GHLA}$};
\end{axis}
  
  \begin{axis}[
    symbolic x coords ={Level 1,Level 2,Level 3,Level 4,Level 5,Level 6,Level 7,Level 8},
    ymin = 0, ymax = 10000,
    hide x axis,
    axis y line*=right,
    yticklabel pos=right,
    ylabel={Average training instances per class},
    ylabel near ticks,
    nodes near coords align={center},
    enlarge x limits=0.1,
    width=12cm,
    height=8.5cm,
  ]
    \addplot[color = black, mark = square*] coordinates {(Level 1,9738) (Level 2,1879) (Level 3,749) (Level 4,540) (Level 5,688) (Level 6,596) (Level 7,636) (Level 8,930)};
  \end{axis}
  
\end{tikzpicture}}
    \caption{NYT Micro-F1.}
    \label{fig:LevelWiseNYTMicro}
    \end{subfigure}
    \hfill
    \begin{subfigure}{0.48\textwidth}
    \centering
    \centerline{\begin{tikzpicture}[scale = 0.58, font = \Large]
  \begin{axis}[
    symbolic x coords ={Level 1,Level 2,Level 3,Level 4,Level 5,Level 6,Level 7,Level 8},
    xticklabel style={font=\small},
    axis y line*=left,
    ybar,
    ylabel={Macro-F1},
    ymin = 40, ymax = 100,
    xlabel near ticks,
    xtick = data,
    ylabel near ticks,
    bar width=0.15cm,
    enlarge x limits=0.1,
    nodes near coords align={center},
    width=12cm,
    height=8.5cm,
  ]
\addplot[draw = black, fill = cyan] 
    coordinates {(Level 1, 87.17) (Level 2, 71.64) (Level 3, 60.57) (Level 4, 61.29) (Level 5, 68.91) (Level 6, 66.21) (Level 7, 55.72) (Level 8, 41.76)};
\addplot[draw = black, fill = lime]  
    coordinates {(Level 1, 87.03) (Level 2, 71.81) (Level 3, 60.74) (Level 4, 62.25) (Level 5, 69.18) (Level 6, 66.75) (Level 7, 56.72) (Level 8, 45.90)};
\addplot[draw = black, fill = orange]  
    coordinates {(Level 1, 87.03) (Level 2, 72.08) (Level 3, 59.91) (Level 4, 61.96) (Level 5, 68.86) (Level 6, 65.96) (Level 7, 57.31) (Level 8, 42.88)};
\addplot[draw = black, fill = pink] 
    coordinates {(Level 1, 87.11) (Level 2, 71.71) (Level 3, 60.53) (Level 4, 62.02) (Level 5, 68.54) (Level 6, 67.18) (Level 7, 56.22) (Level 8, 43.68)};

\legend {AttCNN$_\text{GA}$, AttCNN$_\text{GHLA}$, TopicAttCNN$_\text{GA}$, TopicAttCNN$_\text{GHLA}$};
\end{axis}
  
  \begin{axis}[
    symbolic x coords ={Level 1,Level 2,Level 3,Level 4,Level 5,Level 6,Level 7,Level 8},
    ymin = 0, ymax = 10000,
    hide x axis,
    axis y line*=right,
    yticklabel pos=right,
    ylabel={Average training instances per class},
    ylabel near ticks,
    nodes near coords align={center},
    enlarge x limits=0.1,
    width=12cm,
    height=8.5cm,
  ]
    \addplot[color = black, mark = square*] coordinates {(Level 1,9738) (Level 2,1879) (Level 3,749) (Level 4,540) (Level 5,688) (Level 6,596) (Level 7,636) (Level 8,930)};
  \end{axis}
  
\end{tikzpicture}}
    \caption{NYT Macro-F1.}
    \label{fig:LevelWiseNYTMacro}
    \end{subfigure}
\caption[Level-wise performance results of the AttCNN and TopAttCNN models on the three benchmark datasets.]{Level-wise performance results of the AttCNN and TopAttCNN models on the three benchmark datasets. The bar plot gives the F1 scores (left y-axis) for the different models at each level of the hierarchy while the line plot shows the average training instances for the classes at a particular level (right y-axis).}
\label{fig:LevelWiseResults}
\end{figure}
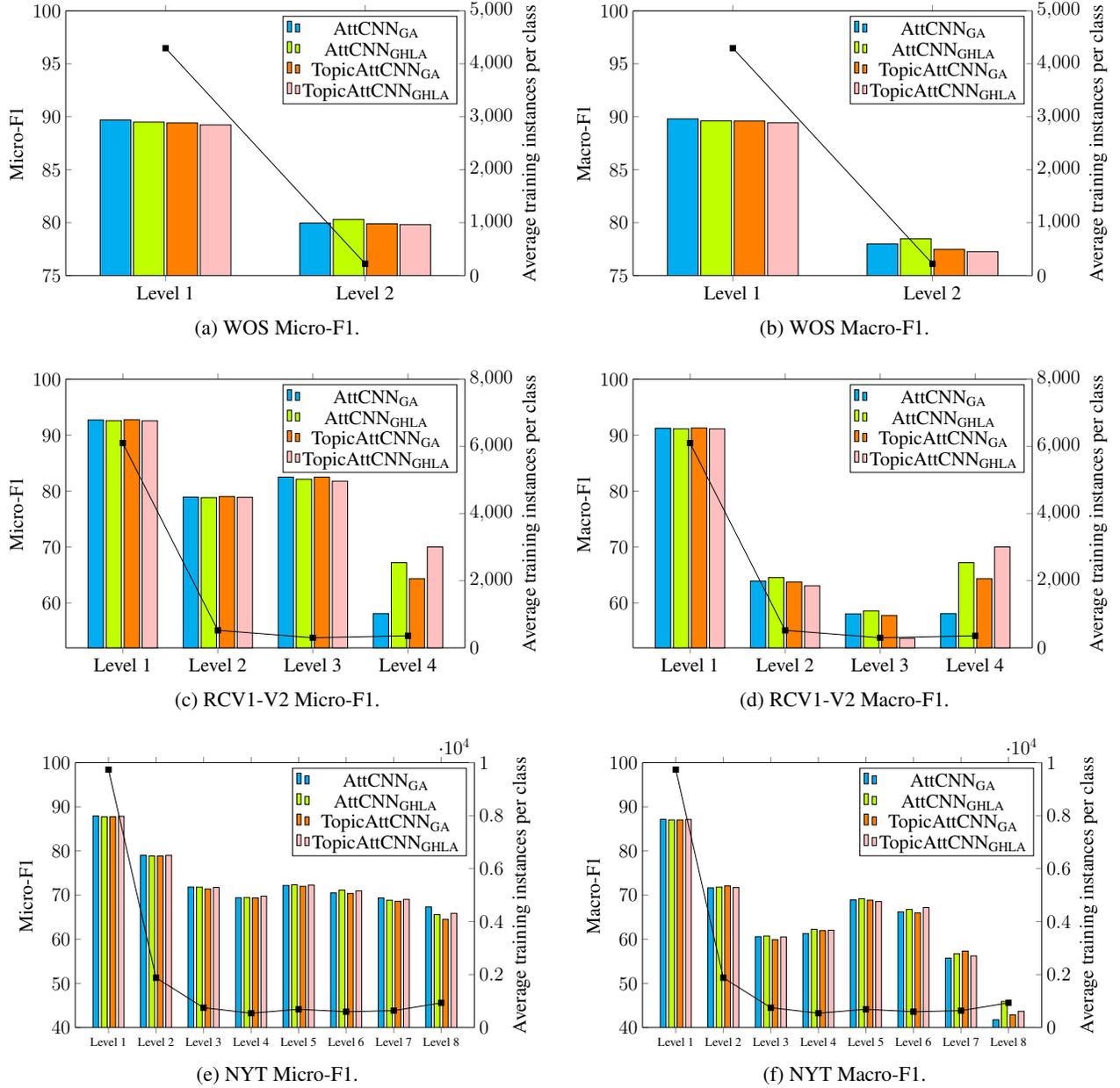

Figure \ref{fig:LevelWiseResults} presents the Micro-F1 and Macro-F1 scores for the classes at each level for the three benchmark datasets. The level-wise results show the general trend that a higher average number of training instances per class at a certain level leads to improved performance at that level. On the WOS dataset we see that the Micro-F1 and Macro-F1 scores are significantly higher for the first-level classes than the second-level classes where the average number of training instances are much larger for level 1 classes (4\,295) than level 2 (224). The performance scores on the RCV1-V2 and NYT dataset also show how the average number of training instances per level generally correlates with the performance differences for the classes at each level. However, the level 8 performance on the NYT dataset shows the worst performance even though it has a higher number of training instances per class than level 3 through 7.

\begin{table}[h]
    \caption[Performance results of the AttCNN and TopAttCNN models under a low-resource scenario.]{Performance results of the AttCNN and TopAttCNN models under a low-resource scenario where only 10\% of the training data is used. For comparison purposes the achieved results when all training data is used are shown in parentheses.}
    \centering
    \label{tab:LowResourceResults}
    
    \scalebox{0.87}{
    
    \begin{tabular}{|l|c|c|c|c|c|c|}
        \hline
        &
        \multicolumn{2}{|c}{\normalsize WOS} &
        \multicolumn{2}{|c}{\normalsize RCV1-V2} &
        \multicolumn{2}{|c|}{\normalsize NYT} \\
        \normalsize Model & {Micro-F1} & {Macro-F1} & {Micro-F1} & {Macro-F1} & {Micro-F1} & {Macro-F1} \\
        \hline
        AttCNN$_\text{GA}$ & 75.30 (84.93) & 61.76 (78.57) & 78.20 (84.67) & 46.75 (62.48) & 70.80 (77.07) & 50.58 (64.08) \\
        AttCNN$_\text{GHLA}$ & \textbf{75.49} (85.00) & \textbf{63.89} (79.02) & 78.32 (84.54) & \textbf{47.44} (63.11) & 71.04 (76.94) & 50.88 (64.57) \\
        TopAttCNN$_\text{GA}$ & 72.37 (84.76) & 57.52 (78.07) & \textbf{78.69} (84.72) & 46.71 (62.33) & \textbf{71.18} (76.88) & \textbf{51.34} (64.18) \\
        TopAttCNN$_\text{GHLA}$ & 73.47 (84.64) & 59.99 (77.86) & 78.35 (84.51) & 45.87 (60.32)  & 70.58 (77.08) & 50.96 (64.35) \\
        \hline
    \end{tabular}
    }
\end{table}

\subsection{Low-resource Results}
Table~\ref{tab:LowResourceResults} shows the performance results of the models under the low-resource scenario where the training data only consists of 10\% of the training set. Performance measures are computed as the average over three runs using different random seeds. The results show that the AttCNN$_\text{GHLA}$ and TopAttCNN$_\text{GA}$ models perform the best on the WOS and NYT datasets respectively. These models also obtain the highest Macro-F1 and Micro-F1 on the RCV1-V2 dataset respectively. However, the TopAttCNN$_\text{GA}$ perform significantly worse than AttCNN$_\text{GHLA}$ on the WOS dataset while the AttCNN$_\text{GHLA}$ model perform similarly to TopAttCNN$_\text{GA}$ on the NYT dataset. Therefore, the results show that AttCNN$_\text{GHLA}$ generally performs the best in the low-resource scenario while AttCNN$_\text{GA}$ performs slightly worse across the three datasets. Furthermore, the results show that the Macro-F1 scores have a much larger decrease than the Micro-F1 scores when moving from using all the training data to the low-resource scenario. This indicates that the Macro-F1 score is a stricter metric since it captures the per-class performance where as the Micro-F1 hides performance information by biasing the results to the performance of the classes with many instances.

\section{Conclusion}
In this paper we evaluate a hierarchical text classification (HTC) approach which uses a pre-trained language model (PLM) and a topic model to extract features from text documents to train a classifier model. The objective of the investigation was to test the claim that the combination of these models improve classification performance on HTC tasks since it has proven to be beneficial for other tasks. The rationale behind the combination of these feature extractor models as proposed by others is that the local semantic and global topic representations, obtained by the PLM and topic model respectively, may provide different granularities of the text documents and therefore allow the classifier to better distinguish between different classes. To evaluate this claim, we use the features obtained from the two models by passing them through separate convolutional layers after which the outputs are combined and used in label-wise attention and classification layers to obtain the final class predictions. We evaluated models using this merging approach with two different label-wise attention mechanisms and compared their performances to a model for which the topic model features were removed. Through comprehensive experiments on three benchmark datasets we showed that the inclusion of the topic model features generally leads to worse performance with the proposed classifier architecture. Therefore, in contrast to previous works, we show that it is not always advisable to incorporate topic model features for classification tasks and that the use of these features should not be assumed beneficial.
%
%
%
\bibliographystyle{unsrtnat}
\bibliography{references}

\begin{thebibliography}{41}
\providecommand{\natexlab}[1]{#1}
\providecommand{\url}[1]{\texttt{#1}}
\expandafter\ifx\csname urlstyle\endcsname\relax
  \providecommand{\doi}[1]{doi: #1}\else
  \providecommand{\doi}{doi: \begingroup \urlstyle{rm}\Url}\fi

\bibitem[Chen et~al.(2021)Chen, Ma, Lin, and Yan]{Chen}
Haibin Chen, Qianli Ma, Zhenxi Lin, and Jiangyue Yan.
\newblock Hierarchy-aware label semantics matching network for hierarchical
  text classification.
\newblock In \emph{Proceedings of the 59th Annual Meeting of the Association
  for Computational Linguistics and the 11th International Joint Conference on
  Natural Language Processing}, pages 4370--4379, Online, 2021. Association for
  Computational Linguistics.
\newblock \doi{10.18653/v1/2021.acl-long.337}.

\bibitem[Huang et~al.(2022)Huang, Liu, Xiao, Zhao, Pan, Zhang, Yang, and
  Liu]{Huang2}
Wei Huang, Chen Liu, Bo~Xiao, Yihua Zhao, Zhaoming Pan, Zhimin Zhang, Xinyun
  Yang, and Guiquan Liu.
\newblock Exploring label hierarchy in a generative way for hierarchical text
  classification.
\newblock In \emph{Proceedings of the 29th International Conference on
  Computational Linguistics}, pages 1116--1127, Gyeongju, Republic of Korea,
  2022. International Committee on Computational Linguistics.

\bibitem[Jiang et~al.(2022)Jiang, Wang, Sun, Chen, Zhuang, and Yang]{Jiang}
Ting Jiang, Deqing Wang, Leilei Sun, Zhongzhi Chen, Fuzhen Zhuang, and Qinghong
  Yang.
\newblock Exploiting global and local hierarchies for hierarchical text
  classification.
\newblock In \emph{Proceedings of the 2022 Conference on Empirical Methods in
  Natural Language Processing}, pages 4030--4039, Abu Dhabi, United Arab
  Emirates, 2022. Association for Computational Linguistics.

\bibitem[du~Toit and Dunaiski(2024)]{dutoit2}
Jaco du~Toit and Marcel Dunaiski.
\newblock Prompt tuning discriminative language models for hierarchical text
  classification.
\newblock \emph{Natural Language Processing}, pages 1--18, 2024.
\newblock \doi{10.1017/nlp.2024.51}.

\bibitem[Wang et~al.(2022{\natexlab{a}})Wang, Wang, Huang, Sun, and
  Wang]{Wang1}
Zihan Wang, Peiyi Wang, Lianzhe Huang, Xin Sun, and Houfeng Wang.
\newblock Incorporating hierarchy into text encoder: a contrastive learning
  approach for hierarchical text classification.
\newblock In \emph{Proceedings of the 60th Annual Meeting of the Association
  for Computational Linguistics}, pages 7109--7119, Dublin, Ireland,
  2022{\natexlab{a}}. Association for Computational Linguistics.
\newblock \doi{10.18653/v1/2022.acl-long.491}.

\bibitem[Wang et~al.(2022{\natexlab{b}})Wang, Wang, Liu, Lin, Cao, Sui, and
  Wang]{Wang2}
Zihan Wang, Peiyi Wang, Tianyu Liu, Binghuai Lin, Yunbo Cao, Zhifang Sui, and
  Houfeng Wang.
\newblock {HPT}: Hierarchy-aware prompt tuning for hierarchical text
  classification.
\newblock In \emph{Proceedings of the 2022 Conference on Empirical Methods in
  Natural Language Processing}, pages 3740--3751, Abu Dhabi, United Arab
  Emirates, 2022{\natexlab{b}}. Association for Computational Linguistics.

\bibitem[Blei(2012)]{topicmodels}
D.M. Blei.
\newblock Probabilistic topic models.
\newblock \emph{Communications of the ACM}, 55\penalty0 (4):\penalty0 77--84,
  2012.
\newblock \doi{10.1145/2133806.2133826}.

\bibitem[Liu et~al.(2021)Liu, Pang, Li, Zhou, and Yue]{liu2021talbert}
W.~Liu, J.~Pang, N.~Li, X.~Zhou, and F.~Yue.
\newblock Research on multi-label text classification method based on
  t{ALBERT}-{CNN}.
\newblock \emph{International Journal of Computational Intelligence Systems},
  14\penalty0 (1), 2021.

\bibitem[Blei et~al.(2003)Blei, Ng, and Jordan]{lda}
D.M. Blei, A.Y. Ng, and M.I. Jordan.
\newblock Latent {D}irichlet {A}llocation.
\newblock \emph{Journal of Machine Learning Research}, 3:\penalty0 993--1022,
  2003.

\bibitem[LeCun et~al.(1989)LeCun, Boser, Denker, Henderson, Howard, Hubbard,
  and Jackel]{cnn}
Yann LeCun, Bernhard Boser, John Denker, Donnie Henderson, R.~Howard, Wayne
  Hubbard, and Lawrence Jackel.
\newblock Handwritten digit recognition with a back-propagation network.
\newblock In \emph{Advances in Neural Information Processing Systems},
  volume~2, pages 396--404, Denver, Colorado, USA, 1989. Morgan-Kaufmann.

\bibitem[Devlin et~al.(2019)Devlin, Chang, Lee, and Toutanova]{Devlin}
Jacob Devlin, Ming-Wei Chang, Kenton Lee, and Kristina Toutanova.
\newblock {BERT}: Pre-training of deep bidirectional transformers for language
  understanding.
\newblock In \emph{Proceedings of the 2019 Conference of the North {A}merican
  Chapter of the Association for Computational Linguistics: Human Language
  Technologies}, pages 4171--4186, Minneapolis, Minnesota, USA, 2019.
  Association for Computational Linguistics.
\newblock \doi{10.18653/v1/N19-1423}.

\bibitem[Grootendorst(2022)]{grootendorst2022bertopic}
Maarten Grootendorst.
\newblock {BERT}opic: Neural topic modeling with a class-based {TF-IDF}
  procedure.
\newblock \emph{arXiv preprint arXiv:2203.05794}, 2022.

\bibitem[Vaswani et~al.(2017)Vaswani, Shazeer, Parmar, Uszkoreit, Jones, Gomez,
  Kaiser, and Polosukhin]{Vaswani}
Ashish Vaswani, Noam Shazeer, Niki Parmar, Jakob Uszkoreit, Llion Jones,
  Aidan~N Gomez, \L~ukasz Kaiser, and Illia Polosukhin.
\newblock Attention is all you need.
\newblock In \emph{Advances in Neural Information Processing Systems},
  volume~30, pages 6000--6010, Long Beach, California, USA, 2017. Curran
  Associates, Inc.

\bibitem[Reimers and Gurevych(2019)]{Reimers2019SentenceBERTSE}
Nils Reimers and Iryna Gurevych.
\newblock Sentence-{BERT}: Sentence embeddings using {S}iamese {BERT}-networks.
\newblock In \emph{Proceedings of the 2019 Conference on Empirical Methods in
  Natural Language Processing and the 9th International Joint Conference on
  Natural Language Processing}, pages 3982--3992, Hong Kong, China, 2019.
  Association for Computational Linguistics.
\newblock \doi{10.18653/v1/D19-1410}.

\bibitem[McInnes et~al.(2018)McInnes, Healy, Saul, and
  Gro{\ss}berger]{McInnes2018UMAP}
Leland McInnes, John Healy, Nathaniel Saul, and Lukas Gro{\ss}berger.
\newblock {UMAP}: Uniform manifold approximation and projection.
\newblock \emph{Journal of Open Source Software}, 3\penalty0 (29):\penalty0
  861, 2018.
\newblock \doi{10.21105/joss.00861}.

\bibitem[McInnes et~al.(2017)McInnes, Healy, and Astels]{McInnes2017HDBSCAN}
Leland McInnes, John Healy, and Steve Astels.
\newblock {HDBSCAN}: Hierarchical density based clustering.
\newblock \emph{Journal of Open Source Software}, 2\penalty0 (11):\penalty0
  205, 2017.
\newblock \doi{10.21105/joss.00205}.

\bibitem[Mullenbach et~al.(2018)Mullenbach, Wiegreffe, Duke, Sun, and
  Eisenstein]{Mullenbach}
James Mullenbach, Sarah Wiegreffe, Jon Duke, Jimeng Sun, and Jacob Eisenstein.
\newblock Explainable prediction of medical codes from clinical text.
\newblock In \emph{Proceedings of the 2018 Conference of the North {A}merican
  Chapter of the Association for Computational Linguistics: Human Language
  Technologies}, pages 1101--1111, New Orleans, Louisiana, USA, 2018.
  Association for Computational Linguistics.
\newblock \doi{10.18653/v1/N18-1100}.

\bibitem[Vu et~al.(2020)Vu, Nguyen, and Nguyen]{Vu}
Thanh Vu, Dat~Quoc Nguyen, and Anthony Nguyen.
\newblock A label attention model for {ICD} coding from clinical text.
\newblock In \emph{Proceedings of the Twenty-Ninth International Joint
  Conference on Artificial Intelligence}, pages 3335--3341, Yokohama, Japan,
  2020.
\newblock \doi{10.24963/ijcai.2020/461}.

\bibitem[Hochreiter and Schmidhuber(1997)]{lstm}
S.~Hochreiter and J.~Schmidhuber.
\newblock Long short-term memory.
\newblock \emph{Neural computation}, 9:\penalty0 1735--1780, 1997.
\newblock \doi{10.1162/neco.1997.9.8.1735}.

\bibitem[Liu et~al.(2022)Liu, Perez-Concha, Nguyen, Bennett, and
  Jorm]{LiuHiLAT}
Leibo Liu, Oscar Perez-Concha, Anthony Nguyen, Vicki Bennett, and Louisa Jorm.
\newblock Hierarchical label-wise attention transformer model for explainable
  {ICD} coding.
\newblock \emph{Journal of Biomedical Informatics}, 133:\penalty0 104161, 2022.
\newblock \doi{https://doi.org/10.1016/j.jbi.2022.104161}.

\bibitem[Yang et~al.(2019)Yang, Dai, Yang, Carbonell, Salakhutdinov, and
  Le]{XLNet}
Zhilin Yang, Zihang Dai, Yiming Yang, Jaime Carbonell, Russ~R Salakhutdinov,
  and Quoc~V Le.
\newblock {XLN}et: Generalized autoregressive pretraining for language
  understanding.
\newblock In \emph{Advances in Neural Information Processing Systems},
  volume~32, pages 5753--5763, Vancouver, Canada, 2019. Curran Associates, Inc.

\bibitem[Banerjee et~al.(2019)Banerjee, Akkaya, Perez-Sorrosal, and
  Tsioutsiouliklis]{Banerjee}
Siddhartha Banerjee, Cem Akkaya, Francisco Perez-Sorrosal, and Kostas
  Tsioutsiouliklis.
\newblock Hierarchical transfer learning for multi-label text classification.
\newblock In \emph{Proceedings of the 57th Annual Meeting of the Association
  for Computational Linguistics}, pages 6295--6300, Florence, Italy, 2019.
  Association for Computational Linguistics.
\newblock \doi{10.18653/v1/P19-1633}.

\bibitem[Koller and Sahami(1997)]{KollerSahami}
Daphne Koller and Mehran Sahami.
\newblock Hierarchically classifying documents using very few words.
\newblock In \emph{Proceedings of the Fourteenth International Conference on
  Machine Learning}, pages 170--178, San Francisco, California, USA, 1997.
  Morgan Kaufmann Publishers Inc.

\bibitem[Kowsari et~al.(2017)Kowsari, Brown, Heidarysafa, Meimandi, Gerber, and
  Barnes]{Kowsari}
Kamran Kowsari, Donald~E Brown, Mojtaba Heidarysafa, Kiana~Jafari Meimandi,
  Matthew~S Gerber, and Laura~E Barnes.
\newblock {HDLT}ex: Hierarchical deep learning for text classification.
\newblock In \emph{16th IEEE International Conference on Machine Learning and
  Applications}, pages 364--371, Cancun, Mexico, 2017.
\newblock \doi{10.1109/ICMLA.2017.0-134}.

\bibitem[Shimura et~al.(2018)Shimura, Li, and Fukumoto]{Shimura}
Kazuya Shimura, Jiyi Li, and Fumiyo Fukumoto.
\newblock {HFT}-{CNN}: Learning hierarchical category structure for multi-label
  short text categorization.
\newblock In \emph{Proceedings of the 2018 Conference on Empirical Methods in
  Natural Language Processing}, pages 811--816, Brussels, Belgium, 2018.
  Association for Computational Linguistics.
\newblock \doi{10.18653/v1/D18-1093}.

\bibitem[du~Toit et~al.(2024)du~Toit, Redelinghuys, and
  Dunaiski]{dutoit2024introducing}
Jaco du~Toit, Herman Redelinghuys, and Marcel Dunaiski.
\newblock Introducing three new benchmark datasets for hierarchical text
  classification, 2024.
\newblock URL \url{https://arxiv.org/abs/2411.19119}.

\bibitem[Gopal and Yang(2013)]{GopalYang}
Siddharth Gopal and Yiming Yang.
\newblock Recursive regularization for large-scale classification with
  hierarchical and graphical dependencies.
\newblock In \emph{Proceedings of the 19th ACM SIGKDD International Conference
  on Knowledge Discovery and Data Mining}, pages 257--265, New York, USA, 2013.
  Association for Computing Machinery.
\newblock \doi{10.1145/2487575.2487644}.

\bibitem[Mao et~al.(2019)Mao, Tian, Han, and Ren]{Mao}
Yuning Mao, Jingjing Tian, Jiawei Han, and Xiang Ren.
\newblock Hierarchical text classification with reinforced label assignment.
\newblock In \emph{Proceedings of the 2019 Conference on Empirical Methods in
  Natural Language Processing and the 9th International Joint Conference on
  Natural Language Processing}, pages 445--455, Hong Kong, China, 2019.
  Association for Computational Linguistics.
\newblock \doi{10.18653/v1/D19-1042}.

\bibitem[Wu et~al.(2019)Wu, Xiong, and Wang]{Wu}
Jiawei Wu, Wenhan Xiong, and William~Yang Wang.
\newblock Learning to learn and predict: A meta-learning approach for
  multi-label classification.
\newblock In \emph{Proceedings of the 2019 Conference on Empirical Methods in
  Natural Language Processing and the 9th International Joint Conference on
  Natural Language Processing}, pages 4354--4364, Hong Kong, China, 2019.
  Association for Computational Linguistics.
\newblock \doi{10.18653/v1/D19-1444}.

\bibitem[Peng et~al.(2021)Peng, Li, Wang, Wang, Gong, Yang, Li, Yu, and
  He]{Peng}
Hao Peng, Jianxin Li, Senzhang Wang, Lihong Wang, Qiran Gong, Renyu Yang,
  Bo~Li, Philip~S. Yu, and Lifang He.
\newblock Hierarchical taxonomy-aware and attentional graph capsule {RCNN}s for
  large-scale multi-label text classification.
\newblock \emph{IEEE Transactions on Knowledge and Data Engineering},
  33\penalty0 (6):\penalty0 2505--2519, 2021.
\newblock \doi{10.1109/TKDE.2019.2959991}.

\bibitem[Deng et~al.(2021)Deng, Peng, He, Li, and Yu]{Deng}
Zhongfen Deng, Hao Peng, Dongxiao He, Jianxin Li, and Philip Yu.
\newblock {HTCI}nfo{M}ax: A global model for hierarchical text classification
  via information maximization.
\newblock In \emph{Proceedings of the 2021 Conference of the North American
  Chapter of the Association for Computational Linguistics: Human Language
  Technologies}, pages 3259--3265, Online, 2021. Association for Computational
  Linguistics.
\newblock \doi{10.18653/v1/2021.naacl-main.260}.

\bibitem[Zhou et~al.(2020)Zhou, Ma, Long, Xu, Ding, Zhang, Xie, and Liu]{Zhou}
Jie Zhou, Chunping Ma, Dingkun Long, Guangwei Xu, Ning Ding, Haoyu Zhang,
  Pengjun Xie, and Gongshen Liu.
\newblock Hierarchy-aware global model for hierarchical text classification.
\newblock In \emph{Proceedings of the 58th Annual Meeting of the Association
  for Computational Linguistics}, pages 1106--1117, Online, 2020. Association
  for Computational Linguistics.
\newblock \doi{10.18653/v1/2020.acl-main.104}.

\bibitem[Raffel et~al.(2020)Raffel, Shazeer, Roberts, Lee, Narang, Matena,
  Zhou, Li, and Liu]{Raffel}
Colin Raffel, Noam Shazeer, Adam Roberts, Katherine Lee, Sharan Narang, Michael
  Matena, Yanqi Zhou, Wei Li, and Peter~J. Liu.
\newblock Exploring the limits of transfer learning with a unified text-to-text
  transformer.
\newblock \emph{Journal of Machine Learning Research}, 21\penalty0
  (1):\penalty0 5485--5551, 2020.

\bibitem[du~Toit and Dunaiski(2023)]{duToitLabelwise}
Jaco du~Toit and Marcel Dunaiski.
\newblock Hierarchical text classification using language models with global
  label-wise attention mechanisms.
\newblock In \emph{Artificial Intelligence Research. SACAIR 2023.
  Communications in Computer and Information Science}, volume 1976, pages
  267--284, 2023.

\bibitem[Clark et~al.(2020)Clark, Luong, Le, and Manning]{electra}
Kevin Clark, Minh{-}Thang Luong, Quoc~V. Le, and Christopher~D. Manning.
\newblock {ELECTRA}: Pre-training text encoders as discriminators rather than
  generators.
\newblock In \emph{Proceedings of the 8th International Conference on Learning
  Representations}, Addis Ababa, Ethiopia, 2020. OpenReview.net.

\bibitem[He et~al.(2021)He, Gao, and Chen]{He2021DeBERTaV3}
Pengcheng He, Jianfeng Gao, and Weizhu Chen.
\newblock {DeBERTaV3}: Improving {DeBERTa} using {ELECTRA-}style pre-training
  with gradient-disentangled embedding sharing.
\newblock \emph{arXiv preprint arXiv:2111.09543}, 2021.

\bibitem[Peinelt et~al.(2020)Peinelt, Nguyen, and Liakata]{peinelt2020tbert}
Nicole Peinelt, Dong Nguyen, and Maria Liakata.
\newblock t{BERT}: Topic models and {BERT} joining forces for semantic
  similarity detection.
\newblock In \emph{Proceedings of the 58th Annual Meeting of the Association
  for Computational Linguistics}, pages 7047--7055, Online, 2020. Association
  for Computational Linguistics.
\newblock \doi{10.18653/v1/2020.acl-main.630}.

\bibitem[Strydom et~al.(2023)Strydom, Dreyer, and van~der Merwe]{StrydomPaper}
Stefan Strydom, Andrei~Michael Dreyer, and Brink van~der Merwe.
\newblock Automatic assignment of diagnosis codes to free-form text medical
  note.
\newblock \emph{Journal of Universal Computer Science}, 29\penalty0
  (4):\penalty0 349--373, 2023.
\newblock \doi{10.3897/jucs.89923}.

\bibitem[Lewis et~al.(2004)Lewis, Yang, Rose, and Li]{RCV1}
David~D. Lewis, Yiming Yang, Tony~G. Rose, and Fan Li.
\newblock {RCV}1: A new benchmark collection for text categorization research.
\newblock \emph{Journal of Machine Learning Research}, 5:\penalty0 361--397,
  2004.

\bibitem[Sandhaus(2008)]{nyt}
Evan Sandhaus.
\newblock The {N}ew {Y}ork {T}imes annotated corpus.
\newblock Technical report, Linguistic Data Consortium, Philadelphia, 2008.

\bibitem[Kingma and Ba(2015)]{kingma2017adam}
Diederik Kingma and Jimmy Ba.
\newblock Adam: {A} method for stochastic optimization.
\newblock In \emph{Proceedings of the 3rd International Conference on Learning
  Representations}, San Diego, California, USA, 2015.

\end{thebibliography}
\end{document}